\documentclass[sigconf, nonacm]{acmart}

\usepackage{hyperref}
\usepackage{amsmath}
\makeatletter
\@ifundefined{Bbbk}{}{%
  
}
\makeatother
\usepackage{amssymb}
\usepackage{mathtools}
\usepackage{amsthm}
\usepackage{multirow}
\usepackage{xcolor}
\usepackage{booktabs}

\usepackage{tikz}
\usepackage[most]{tcolorbox}
\usepackage{enumitem}
\usepackage{textcomp} 

\definecolor{TaskInstruction}{RGB}{198,217,241} 
\definecolor{NodeDescription}{RGB}{255,214,153} 
\definecolor{NodePattern}{RGB}{197,224,180}     
\definecolor{HistoricalInput}{RGB}{216,191,216} 
\definecolor{PredictionToken}{RGB}{190,190,220} 
\definecolor{FutureToken}{RGB}{240,230,140}     

\tcbset{
  vldbPromptBox/.style={
    breakable,
    enhanced,
    colback=white,
    colframe=blue!60!black,
    coltitle=white,
    sharp corners,
    boxrule=0.8pt,
    fonttitle=\bfseries,
    left=1em,right=1em,top=0.9em,bottom=0.9em,
    titlerule=0mm,
    attach boxed title to top left={xshift=0.0em,yshift*=-\tcboxedtitleheight},
    boxed title style={
      sharp corners,
      colback=blue!60!black,
      boxrule=0pt,
      top=0.5ex,bottom=0.5ex,left=0.9em,right=0.9em
    }
  }
}

\begin{document}
\title{STRATA-TS: Selective Knowledge Transfer for Urban Time Series Forecasting with Retrieval-Guided Reasoning}

\author{Yue Jiang}
\affiliation{%
  \institution{Nanyang Technological University}
  \city{Singapore}
}
\email{yue013@e.ntu.edu.sg}

\author{Chenxi Liu}
\affiliation{%
  \institution{Nanyang Technological University}
  \city{Singapore}
}
\email{chenxi.liu@ntu.edu.sg}

\author{Yile Chen}
\affiliation{%
  \institution{Nanyang Technological University}
  \city{Singapore}
}
\email{yile001@e.ntu.edu.sg}

\author{Qin Chao}
\affiliation{%
  \institution{Nanyang Technological University}
  \city{Singapore}
}
\email{chao0009@e.ntu.edu.sg}

\author{Shuai Liu}
\affiliation{%
  \institution{Nanyang Technological University}
  \city{Singapore}
}
\email{shuai004@e.ntu.edu.sg}

\author{Cheng Long}
\affiliation{%
  \institution{Nanyang Technological University}
  \city{Singapore}
}
\email{c.long@ntu.edu.sg}

\author{Gao Cong}
\affiliation{%
  \institution{Nanyang Technological University}
  \city{Singapore}
}
\email{gaocong@ntu.edu.sg}

\begin{abstract}
Urban forecasting models often face a severe \emph{data imbalance problem}: only a few cities have dense, long-span records, while many others expose short or incomplete histories. Direct transfer from data-rich to data-scarce cities is unreliable because only a limited subset of source patterns truly benefits the target domain, whereas indiscriminate transfer risks introducing noise and negative transfer. We present \textbf{STRATA-TS} (Selective TRAnsfer via TArget-aware retrieval for Time Series), a framework that combines domain-adapted retrieval with reasoning-capable large models to improve forecasting in scarce-data regimes. STRATA-TS employs a patch-based temporal encoder to identify source subsequences that are semantically and dynamically aligned with the target query. These retrieved exemplars are then injected into a retrieval-guided reasoning stage, where an LLM performs structured inference over target inputs and retrieved support. To enable efficient deployment, we distill the reasoning process into a compact open model via supervised fine-tuning. Extensive experiments on three parking availability datasets across Singapore, Nottingham, and Glasgow demonstrate that STRATA-TS consistently outperforms strong forecasting and transfer baselines, while providing interpretable knowledge transfer pathways.
\end{abstract}

\maketitle

\section{Introduction}
Time series forecasting underpins a wide range of urban computing tasks, including traffic flow prediction, parking demand estimation, energy scheduling, and mobility optimization. Advances in deep learning—particularly spatio-temporal graph neural networks (GNNs) \cite{DCRNN,GTS,AGCRN,MTGNN,STGCN} and sequence-based architectures \cite{Autoformer,informer,itransformer}—have significantly improved the modeling of temporal dependencies and spatial interactions. Yet, the widespread deployment of these methods is constrained by the availability of long-term, high-quality data. Many cities face data scarcity due to sparse sensing infrastructures, privacy considerations, or the early stage of system roll-out. Scarcity manifests both spatially (e.g., too few monitored sites) and temporally (e.g., limited historical coverage), severely impairing model generalization.

A promising remedy is to leverage data-rich cities to assist data-poor ones, i.e., cross-city transfer learning. The idea is to adapt useful knowledge from source cities so that models perform well in target cities with minimal labeled data. Previous research has explored pretraining-finetuning pipelines, common embedding spaces, and transferable graph structures \cite{CrossTReS,RegionTrans,TransGTR,ST-GFS}. While feasible, these methods face two fundamental issues. First, they often overlook the heterogeneity across cities—differences in road networks, land use, demand rhythms, and seasonality—which can lead to mismatched transfers and suboptimal forecasts. Second, most methods rely on parameter adaptation in a black-box manner, which limits interpretability and fails to incorporate higher-level reasoning, reducing robustness in practical deployments.

\begin{figure}[t]
  \centering
  \includegraphics[width=0.92\linewidth]{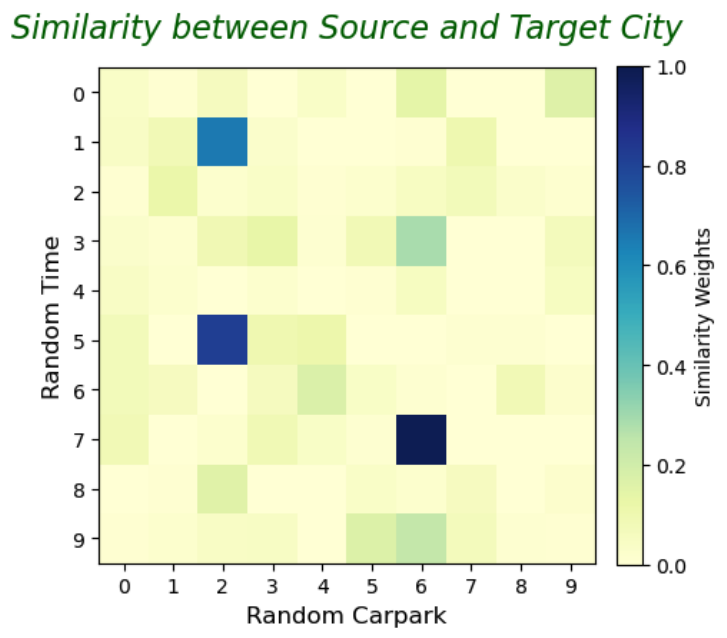}
  \caption{Similarity heatmap illustrating the example above. Each cell shows the similarity score $w_i$ in Eq.~\eqref{eq:sim_weight} between a Nottingham \emph{3-hour} target slice and a randomly sampled Singapore slice (carpark on the $x$-axis, time window on the $y$-axis). Fewer than $5\%$ of the samples achieve $w_i>0.5$, indicating that only a small subset of source subsequences are truly relevant and motivating the need for \emph{target-aware selective retrieval}.}
  \label{fig:intro_similarity_heatmap}
\end{figure}

In reality, urban patterns vary dramatically across contexts. For example, demand around a business district in Singapore can be governed by very different rhythms compared to residential facilities in Nottingham. As illustrated in Fig.~\ref{fig:intro_similarity_heatmap}, only a very small portion (fewer than 5\%) of source subsequences from Singapore display strong similarity ($w_i > 0.5$) with a Nottingham query slice. This observation emphasizes that most source information is irrelevant or even misleading, and naive transfer may cause negative transfer. Furthermore, the correlation structures among locations differ substantially across cities, leading to divergent graph topologies that full-domain transfer methods fail to account for.

To overcome these challenges, we propose \textbf{STRATA-TS}, a retrieval- and reasoning-enhanced framework for urban time series forecasting. The framework is built upon two main components: a \emph{Target-Aware Retriever}, which identifies semantically and temporally aligned source subsequences for each target query, and an \emph{LLM-based Reasoning Module}, which integrates retrieved context with target inputs to generate predictions. Rather than transferring entire source domains, STRATA-TS focuses on targeted retrieval of relevant slices using embeddings obtained from a patch-based temporal encoder trained in a masked autoencoding fashion. These embeddings allow for fine-grained similarity matching and the selection of only informative evidence.

The reasoning stage distinguishes STRATA-TS from prior transfer frameworks. By structuring retrieved subsequences and target context into prompts, we enable large language models to perform in-context reasoning. Specifically, we employ a strong pretrained model (e.g., GPT-o1) to analyze correlations between retrieved evidence and target dynamics, then distill its reasoning into a smaller LLaMA-3-8B backbone via supervised fine-tuning. This approach provides both practical efficiency and improved transparency compared to conventional black-box adaptation.

Importantly, STRATA-TS is designed to be modular and plug-and-play. It can complement existing transfer frameworks such as TransGTR by injecting retrieval-informed reasoning at inference time. We validate our approach on three real-world parking datasets from Singapore, Nottingham, and Glasgow. Results show that STRATA-TS delivers substantial improvements over both traditional forecasting models and recent transfer-learning baselines. Beyond raw accuracy, we also present ablation studies and visualizations to demonstrate how retrieval precision, prompt design, and model adaptation influence outcomes.

In summary, our contributions are as follows:
\begin{itemize}
    \item We introduce STRATA-TS, a framework that combines target-aware retrieval with LLM-based reasoning for selective, interpretable cross-city forecasting.
    \item We design a patch-based temporal encoder for embedding and retrieval, ensuring that only semantically aligned and temporally relevant subsequences are transferred.
    \item We develop a retrieval-guided reasoning stage that distills the capabilities of a large proprietary model into a compact open-source LLM, balancing accuracy, interpretability, and efficiency.
    \item We demonstrate through extensive experiments on three urban parking datasets that STRATA-TS consistently surpasses state-of-the-art baselines and transfer methods.
\end{itemize}

\section{Related Work}
\subsection{Transfer Learning Models}


Time series forecasting has progressed from classical models like ARIMA \cite{ARIMA} to neural approaches such as RNNs and Transformers, which better capture non-linear and long-range dependencies. More recently, spatio-temporal graph neural networks (STGNNs) \cite{AGCRN,DCRNN,TGCN,GTS,D2STGNN} have advanced the field by jointly modeling spatial and temporal correlations. However, their deployment is often hindered by limited data availability in many cities, where collecting sufficient historical records requires long-term sensing infrastructure. To mitigate this, transfer learning has emerged as an effective strategy to transfer knowledge from data-rich source cities to data-scarce target cities.

Early approaches \cite{CrossTReS, MetaST, RegionTrans} focus on grid-based cross-city transfer learning. For instance, RegionTrans \cite{RegionTrans} employs ConvLSTM \cite{ConvLSTM} layers as the backbone and utilizes auxiliary data similarity to guide the transfer process from the source to the target city. CrossTReS \cite{CrossTReS} further refines transfer granularity by decoupling node-level and edge-level knowledge and introducing a weighting network to match regions across cities during training. However, these methods still involve transferring information from all source regions, including those assigned low relevance weights, which may introduce noise and reduce transfer effectiveness. These methods also neglect the temporal dynamics and fail to retrieve the most relevant time slices and information.

Recent models \cite{DASTNet,ST-GFS,TransGTR} extend transfer learning to graph-structured urban data by leveraging techniques like meta-learning and transferable graph structure adaptation. For example, ST-GFSL \cite{ST-GFS} employs meta-learning to facilitate knowledge transfer across spatio-temporal graphs, while TransGTR \cite{TransGTR} introduces a structure learning framework that learns a transferable graph constructor from the source city and applies it to the target city. Despite these advancements, such methods generally perform full-domain transfer using pre-defined or globally learned graphs, without explicitly modeling the heterogeneous relevance between source and target nodes. As a result, they may incorporate irrelevant or noisy source patterns, leading to suboptimal forecasting performance and reduced transfer effectiveness.


\subsection{Large Language Models}

Recent advancements in large language models (LLMs) \cite{llm0shot,llmfewshot,fromnews} have expanded their applicability beyond traditional natural language processing tasks, demonstrating growing potential in time series forecasting, particularly in transfer learning and few-shot learning settings. A key strength of LLMs lies in their ability to adapt pretrained knowledge to new tasks with limited data, making them well-suited for cross-city forecasting scenarios where target domains often suffer from data scarcity and effective knowledge transfer is crucial. Several pioneering studies \cite{Timellm, GPT4TS, llm4ts} have explored adapting LLMs for time series applications. For example, GPT4TS \cite{GPT4TS} and LLM4TS \cite{llm4ts} repurpose language modeling architectures such as GPT-2 by introducing numerical tokenization and structured fine-tuning strategies, yielding moderate forecasting performance. Time-LLM \cite{Timellm} further advances this line of work by reprogramming temporal inputs into prompt-style formats and applying the Prompt-as-Prefix approach to enable autoregressive forecasting with LLMs. However, these methods typically overlook the spatial dependencies inherent in multivariate time series and do not incorporate knowledge retrieval and transfer to guide transfer across cities. Additionally, their reliance on purely numerical tokenization limits the model's ability to perform causal reasoning, which is essential for addressing complex, real-world forecasting scenarios.



\subsection{Retrieval-Augmented Generation Enhanced Models}

While large language models (LLMs) have demonstrated strong performance across various forecasting tasks, their generalization often deteriorates in the presence of non-stationary distributions, domain shifts, or limited training data, which are common conditions encountered in cross-city forecasting. Retrieval-Augmented Generation (RAG) provides a promising solution to these challenges by incorporating additional, task-relevant knowledge directly into the inference process, thereby enhancing robustness and adaptability. Recent studies \cite{RATD, RAFT, RAF, TimeRAG, TS-RAG} have begun extending RAG techniques to time series domains. For example, TimeRAG \cite{TimeRAG} retrieves time slices with high temporal similarity between the source sequence and the target sequence directly and incorporates them into a frozen LLM through a reprogramming module to guide sequence prediction. Additionally, RAF \cite{RAF} and TS-RAG \cite{TS-RAG} employ parameter-frozen, pretrained encoders to embed both the query and source time series, and perform retrieval using Euclidean distances in the representation space. These models then concatenate retrieved time series with the input, then feed it to the forecasting models. Despite these advances, existing methods often rely on shallow similarity metrics or domain-agnostic embeddings from a parameter-frozen encoder, without explicitly modeling domain-specific dynamics. As a result, the retrieved sequences may include semantically or dynamically misaligned patterns, leading to suboptimal retrieval precision and reduced forecasting performance.



\section{Method}
\begin{figure*}[ht]
  \centering
  \includegraphics[width=\linewidth]{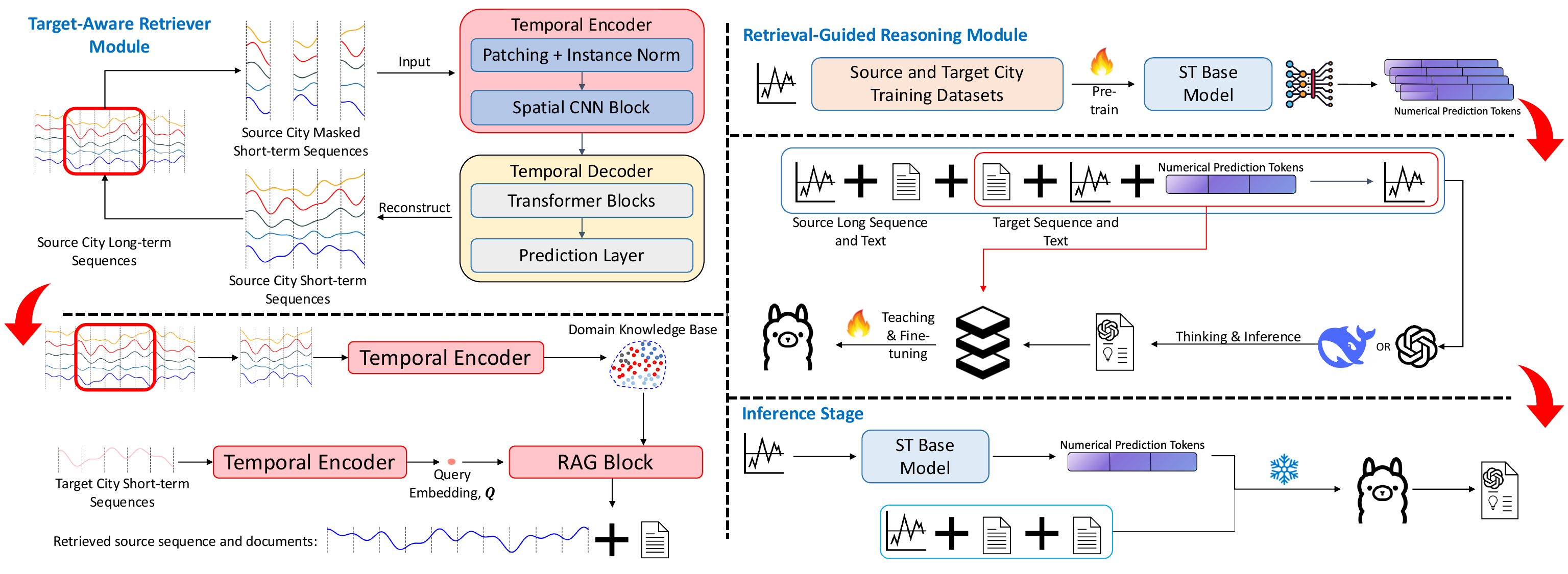}
  \caption{The overall architecture of the proposed STRATA-TS framework. The Target-Aware Retriever Module is on the left, and the Retrieval-Guided Reasoning Module is on the right.}
  \label{fig:STRATA-TS}
\end{figure*}

Our proposed framework, STRATA-TS, is designed to enable selective, precise, and interpretable knowledge transfer for cross-city time series forecasting. The overall framework of STRATA-TS is illustrated in Fig. \ref{fig:STRATA-TS}. Unlike conventional transfer learning approaches that indiscriminately transfer global models or entire graph structures, STRATA-TS focuses on identifying and leveraging only the most semantically relevant source patterns to deliver interpretable forecasting in the target city. To achieve this, STRATA-TS comprises two key components: a Target-Aware Retriever Module, which retrieves source information with high semantic relevance to the target query, and an Retrieval-Guided Reasoning Module, which performs structured reasoning over the retrieved source context and target sequence to generate accurate and explainable predictions. Specifically, we construct paired samples between the source city long-term time sequences (from one week to several months, depending on the source city data spanning) and their associated contextual information. We pretain a patch-based temporal encoder on the source city time sequence in the fashion of the Marked Autoencoder\cite{STEP}. Given a target city query, the model encodes the observed sequence and retrieves top-k source sequences by computing embedding similarity against the paired source set. The corresponding contextual information and the long-term sequence information from these matched source samples are then injected into the Retrieval-Guided Reasoning Module. In the Retrieval-Guided Reasoning Module, we first employ a powerful foundation model such as GPT-o1 or DeepSeekR1 to perform structured reasoning and inference over the retrieved source information, the target contextual and temporal input, and the associated ground-truth label. This process enables the model to disentangle useful domain knowledge from irrelevant signals. We then construct a supervised fine-tuning (SFT) prompt that integrates the retrieved knowledge with the target city's input, and fine-tune an LLaMA-3-8B model to adapt the reasoning patterns and enhance forecasting performance. This fine-tuned model operates efficiently at the inference stage while maintaining strong generalization and interpretability. Together, these modules facilitate targeted knowledge injection and in-context forecasting, enabling robust generalization.

\subsection{Target-Aware Retriever Module}
\label{sec:retriever}

In this module, we design a patch-based Masked Autoencoder (MAE) pre-training pipeline to learn domain-adaptable temporal representations from data-rich source cities, enabling selective retrieval of transferable patterns. The source city dataset typically contains long-term sequences spanning weeks or months. For each retrieval query, we slice from these long-term sequences a short-term segment whose temporal length matches the target city training span (e.g., the latest three days), ensuring a fair and contextually aligned comparison between domains. These short-term sequences are then processed as follows.

Let $\mathbf{X}\in\mathbb{R}^{B\times N\times L}$ denote a mini-batch of multivariate time series, with $B$ samples, $N$ nodes (e.g., car parks), and length $L$. Each sample is segmented into non-overlapping patches of width $p$, producing $L'=L/p$ patches:
\begin{equation}
\mathbf{P}\in\mathbb{R}^{B\times N\times L'\times p},\qquad 
\mathbf{P}_{b,n,t,:}=\mathbf{X}_{b,n,(t-1)p+1:tp}.
\label{eq:patching}
\end{equation}

A linear projection maps each patch to $d$ dimensions, yielding $\mathbf{E}\in\mathbb{R}^{B\times N\times L'\times d}$, followed by instance normalization to mitigate scale heterogeneity:
\begin{equation}
\widehat{\mathbf{E}}=\frac{\mathbf{E}-\mu(\mathbf{E})}{\sigma(\mathbf{E})+\epsilon}.
\label{eq:instnorm}
\end{equation}

We process each node independently by folding the node axis into the batch and arranging the \emph{patch} index and \emph{embedding} dimension as a $2$D grid. Concretely, we reshape
\begin{equation}
\widehat{\mathbf{E}}\in\mathbb{R}^{(B\!\cdot\!N)\times 1 \times L'\times d},
\label{eq:reshape}
\end{equation}
so that height $H=L'$ is the number of patches and width $W=d$ is the hidden dimension. We then apply a spatial CNN with padding to preserve $(H,W)$:
\begin{equation}
\mathbf{Z}'=\underbrace{\mathrm{Conv}^{1\rightarrow 16}_{3\times 3}\!\rightarrow\!\mathrm{ReLU}\!\rightarrow\!
\mathrm{Conv}^{16\rightarrow 16}_{3\times 3}\!\rightarrow\!\mathrm{ReLU}\!\rightarrow\!
\mathrm{Conv}^{16\rightarrow 1}_{1\times 1}}_{\text{Spatial CNN}}(\widehat{\mathbf{E}}).
\label{eq:cnn}
\end{equation}

Finally, we reshape the output back $\widetilde{\mathbf{E}}\in\mathbb{R}^{B\times N\times L'\times d}$ and feed to a stack of Transformer decoder layers. During MAE pre-training, a random subset $\mathcal{M}\subset\{1,\dots,L'\}$ of patches is masked; the encoder sees only visible tokens, and a lightweight decoder reconstructs masked patches back in the original time sequences. Let $\widehat{\mathbf{X}}$ be the reconstruction; the loss is
\begin{equation}
\mathcal{L}_{\mathrm{MAE}}=\big\|\mathbf{M}\odot(\widehat{\mathbf{X}}-\mathbf{X})\big\|_{1},
\label{eq:mae_loss}
\end{equation}
where $\mathbf{M}$ selects masked positions. The decoder is intentionally kept lightweight, as it is used only during MAE pre-training and is discarded thereafter; this encourages the encoder to capture transferable temporal structure.

\noindent\textbf{Top-$K$ Retrieval.}  
After pre-training, each sliced source segment is embedded by the temporal encoder to build a large-scale domain knowledge base
\begin{equation}
\mathcal{D}=\{(\mathrm{id},\mathbf{V},\mathbf{S},T)\mid \mathrm{id}\in\mathcal{I}\},
\label{eq:knowledge_base}
\end{equation}
where $\mathrm{id}$ denotes the identifier of a specific source segment and $\mathcal{I}$ is the full set of segment identifiers in the knowledge base. In our setting, $\mathcal{D}$ contains more than 500k segment samples collected from the source city. Each entry stores $\mathbf{V}\in\mathbb{R}^{L' \times d}$, the representation embedding of segment $\mathrm{id}$, $\mathbf{S}$, the corresponding long-term sequence, and $T$, the associated contextual information. 

Given a target query, the encoder produces the query embedding $\mathbf{Q} \in \mathbb{R}^{L' \times d}$. To improve retrieval precision, we compute similarity as the average of two metrics: the negative L2 distance and the negative Mahalanobis distance, both computed after flattening the matrices to vectors in $\mathbb{R}^{L'd}$. Let 
$\mathbf{q}=\mathrm{vec}(\mathbf{Q}) \in \mathbb{R}^{L'd}$ and $\mathbf{v}=\mathrm{vec}(\mathbf{V}) \in \mathbb{R}^{L'd}$. The L2 distance is:
\begin{equation}
D_{\mathrm{L2}}(\mathbf{q},\mathbf{v}) = \sqrt{\sum_{i=1}^{L'd} (q_i - v_i)^2},
\label{eq:l2}
\end{equation}
and the Mahalanobis distance is:
\begin{equation}
D_{\mathrm{Mah}}(\mathbf{q},\mathbf{v}) = \sqrt{ (\mathbf{q} - \mathbf{v})^\top \mathbf{\Sigma}^{-1} (\mathbf{q} - \mathbf{v}) },
\label{eq:mahalanobis}
\end{equation}
where $\mathbf{\Sigma}$ is the covariance matrix estimated once from all flattened embeddings in $\mathcal{D}$. 

The final similarity score is:
\begin{equation}
\mathrm{sim}(\mathbf{Q},\mathbf{V}) = -\tfrac{1}{2} \left[ D_{\mathrm{L2}}(\mathbf{q},\mathbf{v}) + D_{\mathrm{Mah}}(\mathbf{q},\mathbf{v}) \right].
\label{eq:similarity}
\end{equation}

Additionally, we define a normalized similarity weight for interpretability:
\begin{equation}
w_i \;=\; \frac{\displaystyle \min_{j}\,\big\|\mathbf{x}_j-\mathbf{x}_t\big\|_2}{\displaystyle \big\|\mathbf{s}_i-\mathbf{x}_t\big\|_2}\,,
\label{eq:sim_weight}
\end{equation}
where $\mathbf{s}_i$ is a source slice and $\mathbf{x}_t$ the target slice. This quantifies how much closer $\mathbf{s}_i$ is relative to the best-matching source.

We then retrieve the top-$K$ most relevant segments:
\begin{equation}
\mathcal{N}_K(\mathbf{Q})=\operatorname*{arg\,topK}_{(\mathrm{id},\mathbf{V},\mathbf{S},T)\in\mathcal{D}}\ \mathrm{sim}(\mathbf{Q},\mathbf{V}),
\label{eq:topk}
\end{equation}
and use their $(\mathbf{S},T)$ pairs as structured auxiliary context for the downstream LLM inference.

\subsection{Retrieval-Guided Reasoning Module}
\label{sec:llm-inference}

The Retrieval-Guided Reasoning Module integrates the predictions of a spatio-temporal (ST) base forecasting model with retrieved source-domain context to perform causal reasoning over both textual and numerical modalities. This design enables context-aware correction of base-model outputs while preserving interpretability.

\textbf{Pre-training of the Base ST Model.}
We first train a spatio-temporal base forecasting model using the combined training datasets from the source and target cities. The base model can be instantiated as either a classical time-series forecasting baseline or a transfer-learning backbone. In our implementation, we adopt \textbf{TransGTR}~\cite{TransGTR} due to its strong cross-city forecasting performance. Once trained, the model generates numerical forecasts for \emph{all} training and test samples in the target city. The resulting \emph{prediction tokens} are organized as
\[
\mathcal{C}_{\mathrm{train}} \in \mathbb{R}^{N_{\mathrm{train}} \times N_{\mathrm{nodes}} \times L_{\mathrm{pred}}}, \quad
\mathcal{C}_{\mathrm{test}} \in \mathbb{R}^{N_{\mathrm{test}} \times N_{\mathrm{nodes}} \times L_{\mathrm{pred}}},
\]
where $N_{\mathrm{train}}$/$N_{\mathrm{test}}$ are the number of training/testing sequences, $N_{\mathrm{nodes}}$ is the number of target nodes (e.g., car parks), and $L_{\mathrm{pred}}$ is the forecasting horizon. Each token thus encodes the full $L_{\mathrm{pred}}$-step prediction for a specific node in a specific sample, forming a structured numerical representation of the base model's output.

\textbf{Correlation Extraction via Powerful LLMs.}
To capture high-quality reasoning patterns while minimizing token consumption, we introduce a low-cost correlation extraction stage. For each target-city car park, five representative training instances are sampled. For each selected instance, we construct a structured prompt that concatenates: (1) source-domain context (textual description and the retrieved long-term source sequence); (2) target-domain context (textual description and the observed short-term target sequence); (3) the corresponding prediction tokens from $\mathcal{C}_{\mathrm{train}}$ for the node relevant to that instance; and (4) the ground-truth target sequence for the forecast horizon. This prompt is fed into a reasoning-capable LLM such as GPT-o1 or DeepSeek-R1, which performs causal reasoning to infer correlations between textual semantics, numerical forecasts, and observed ground truth. The model returns structured \emph{hints} that distill these correlations. A prompt example is shown for better illustration.

\begin{tcolorbox}[vldbPromptBox,title={STRATA-TS Reasoning Prompt}]
\small
\textbf{\texttt{[INST]}} 
\textbf{Role:} You are an AI agent specialized in \emph{cross-city transfer learning analysis} for parking availability.  
\textbf{Objective:} Do \emph{not} output future parking values. Instead, use reasoning and causal inference to extract \textbf{useful, generalizable hints} that relate the inputs (textual context, source long-term sequence, target short-term history, and simulation predictions) to the \textbf{ground truth} sequence. These hints will be used later to fine-tune a student model (e.g., LLaMA-3-8B). Focus on: (i) bias/scale calibration of simulation predictions, (ii) lag/seasonality/day-of-week effects, (iii) cross-city pattern alignment.

\medskip
\textbf{Input Data:}
\begin{enumerate}[label=(\arabic*), leftmargin=2.0em, itemsep=0.25em]
  \item \textbf{Source city textual information:} \emph{Jem Shopping Mall, Singapore} — suburban mall in Jurong East, directly connected to Jurong East MRT interchange and adjacent to Westgate. 241 shops across six levels ($\sim$818{,}000 sq ft). Major retailers include IKEA, FairPrice Xtra, H\&M, Uniqlo, Muji, Courts, Don Don Donki. Carpark: 669 bays (B2 \& B3), EV charging at B3. Pricing (incl.\ 9\% GST from 2024-01-01): Mon--Thu first hour \$2.18, subsequent 15 mins \$0.55; Fri--Sun first hour \$2.73, subsequent 15 mins \$0.65; after 6\,pm \$3.27 per entry.
  \item \textbf{Source city long-term sequence:} Retrieved multi-week availability series with high semantic similarity to the target carpark: \texttt{<source\_seq\_values>}.
  \item \textbf{Prediction horizon:} Given 3 hours of history \texttt{2017-01-24 14:30} to \texttt{2017-01-24 17:30} (Tuesday), analyze for the period \texttt{2017-01-24 17:45} to \texttt{2017-01-24 20:30}.
  \item \textbf{Target city textual information:} \emph{St Marks Place NCP Car Park, Newark} — A spacious and affordable multi-story facility with 512 spaces, located on Lombard Street (NG24 1XT) in Newark and Sherwood. Height restriction: 1.98\,m; includes dedicated disabled bays. Open 24/7 with flexible, budget-friendly pricing: \pounds1.95 (1\,h), \pounds3.90 (2\,h), \pounds6.95 (24\,h); early-bird discount before 9{:}00\,am. Payment via ticket, card, phone, or NCP App (Location ID 32089). Known for safety (Parking Safety Award), though navigating inside can be tight due to high concrete sides. Typical patterns: peak availability 8\,pm--6\,am, lowest availability 10\,am--3\,pm.
  \item \textbf{Target city historical records (last 3h):} \texttt{<target\_seq\_values>}.
  \item \textbf{Simulation predictions (base ST model, e.g., TransGTR):} \texttt{<Prediction\_tokens>}.
\end{enumerate}

\medskip
\textbf{Training-time supervision:} The ground truth for the horizon is provided as \texttt{<ground truth>}. Use it \emph{only} to infer relationships (e.g., bias/scale, residual directions, regime shifts). \textbf{Do not} echo the ground truth or produce numeric forecasts.

\medskip
\textbf{Analysis Goals (what to extract):}
\begin{itemize}[leftmargin=1.5em, itemsep=0.2em]
  \item \textbf{Cross-city alignment:} Which parts of the source long-term sequence best align with the target history (e.g., day-of-week, hour-of-day, shopping-peak proximity, transit adjacency)? Provide a short rationale.
  \item \textbf{Lag/seasonality cues:} Dominant lags (e.g., 1, 2, 4 steps), diurnal phase, weekday/weekend effects, and expected monotonicity segments over the horizon.
  \item \textbf{Change-points/regimes:} Any shift boundaries (e.g., pre-/post-evening peak) and how they affect corrections.
\end{itemize}

\medskip
\textbf{Rules:} 
\begin{enumerate}[label=(R\arabic*), leftmargin=2.0em, itemsep=0.2em]
  \item Do \emph{not} output any numeric forecasts or restate \texttt{<ground truth>}.
  \item Reason causally from text + sequences; prefer explanations tied to retail/transit factors and diurnal/weekday structure.
\end{enumerate}
\textbf{\texttt{[/INST]}}
\end{tcolorbox}

\begin{tcolorbox}[vldbPromptBox,title={STRATA-TS: GPT-o1 Response}]
\small
The two carparks share \textbf{similar urban contexts} — both located near high-footfall commercial areas and public transit hubs, leading to \textbf{comparable weekday commuter-driven demand peaks} and \textbf{evening recoveries}. By analyzing the \textbf{long-term weekly patterns from the Singapore source city}, we can infer \textbf{weekend usage profiles absent in the Newark short history}, revealing that \textbf{weekend midday periods are moderately busy but less congested than weekdays}. After \textbf{normalizing and fitting both the source and target sequences}, we \textbf{extract useful patterns} and \textbf{rescale to the target city scale}.  

Based on the given parking lot records and a detailed observation of the available parking spaces over the week, here is a more specific breakdown of when peaks (maximum availability) and dips (minimum availability) occur:  
\textit{On Weekdays (Monday to Friday):} \textbf{Peak availability: From 8 PM to 6 AM} (reaching the maximum capacity of 512 spaces). This reflects the time when parking demand is lowest, typically during the late-night and early morning hours. \textbf{Minimum availability: 10 AM to 3 PM} (223--374 spaces), indicating busiest hours likely due to work or daily activities.  
\textit{On Weekends (Saturday and Sunday):} \textbf{Peak availability: From 8 PM to 6 AM} (full capacity), indicating lower demand during these hours. \textbf{Minimum availability: 11 AM to 4 PM} (248--374 spaces), showing moderate to high usage during midday hours.  
In summary, the parking lot \textbf{consistently reaches full capacity during late evening and early morning hours}, with the \textbf{highest usage during mid-morning to mid-afternoon} across both weekdays and weekends.  

This \textbf{cross-city pattern alignment} suggests that \textbf{source-derived weekend profiles can calibrate weekday-only target observations}, guiding \textbf{scale-adjusted corrections} in the simulation predictions.
\end{tcolorbox}

\textbf{Supervised Fine-tuning of a Lightweight LLM.}
The final goal of the reasoning extraction stage is to distill the high-quality causal inferences from a proprietary reasoning LLM (e.g., GPT-o1) into a smaller, deployable model such as \textbf{LLaMA-3-8B}. This is achieved via supervised fine-tuning (SFT). 

Each \emph{SFT instance} corresponds to a single target-city forecasting case and is constructed as follows:
\begin{enumerate}[leftmargin=2.0em, itemsep=0.25em]
    \item \textbf{Context Encoding:} The prompt contains (i) the target city's textual description; (ii) the short-term observed target sequence; and (iii) the base ST model's prediction tokens from $\mathcal{C}_{\mathrm{train}}$ containing $L_{\mathrm{pred}}$ forecast steps.
    \item \textbf{Reasoning Supervision:} The GPT-o1 output that encodes patterns such as regime-shift boundaries, seasonal alignment cues, and lag dependencies.
    \item \textbf{Ground-truth Alignment:} The ground truth values are expressed in the textual domain, non-numeric format. This ensures the student model learns \emph{reasoning strategies}, not numeric memorization.
\end{enumerate}

By training on this dataset, the lightweight LLaMA-3-8B learns to replicate GPT-o1's reasoning behavior when given new inputs at test time. This SFT process has several advantages: (i) it significantly reduces inference cost compared to running GPT-o1 online; (ii) it considers global temporal dynamics in the target city through batched fine-tuning instead of only focusing on single temporal query, thereby, providing better forecasting performance; and (iii) it preserves interpretability, as the fine-tuned model can be prompted to generate human-readable reasoning traces.

\textbf{Inference Procedure.}
At inference, the ground-truth future sequence is masked. The pre-trained ST base model processes the observed input sequence from the target-city test set to produce the corresponding prediction tokens $\mathcal{C}_{\mathrm{test}}$ (covering all nodes and all $L_{\mathrm{pred}}$ steps per node). These tokens, together with the target city’s context and any retrieved source sequences and metadata, are provided to the fine-tuned LLaMA-3-8B, which performs structured reasoning to yield the final forecast.

This two-stage design (i) separates numerical pattern extraction (ST base model) from high-level reasoning (LLM), improving modularity and interpretability; (ii) distills GPT-o1 style reasoning into a smaller model for efficient deployment; and (iii) unifies numerical predictions, temporal sequences, and semantic descriptions in a single inference pipeline, enabling context-aware correction of base-model errors.

\begin{table*}[hbt!]
    \centering 
    \caption{Statistics of 
    real-world datasets.}\label{tab:Datasetss}
    \begin{tabular}{cccc}  
        \toprule
        Datasets&Time Frequency&Time range&Total Parking Records\\
        \midrule 
         Singapore & 15 mins & 1 May 2021 - 6 June 2021&813,984\\
         Nottingham & 15 mins & 26 Oct 2016 - 16 Feb 2017 & 119,108\\
        Glasgow & 15 mins & 26 Oct 2016 - 16 Feb 2017 & 119,108\\
        \bottomrule 
    \end{tabular}
\end{table*}

\begin{table*}[!htbp] 
    \centering  
    \caption{120 mins Performance comparison on Nottingham dataset trained on SG dataset. Best results in \textcolor{red}{red}, second best in \textcolor{blue}{blue}.}\label{tab:not}
    \begin{tabular}{c|c c c|c c c|c c c|c c c}  
        \hline 
          \multirow{2}*{\rotatebox{90}{ }}&  & 15 mins &  &  & 30 mins &  &  & 45 mins & & & 60 mins & \\ 
        &MAE& RMSE & MAPE &MAE& RMSE & MAPE&MAE& RMSE & MAPE&MAE& RMSE & MAPE\\ 
        \hline 
         DLinear   & 10.58 & 29.47 & 5.37\% & 14.62 & 41.75 & 9.19\% & 20.51 & 51.77 & 13.15\% & 26.85 & 60.53 & 17.40\% \\
         PatchTST  & 16.39 & 43.70 & 14.59\% & 21.85 & 54.15 & 19.19\% & 27.68 & 63.60 & 22.52\% & 32.66 & 71.75 & 27.25\% \\
         GPT4TS    & 9.17  & 30.31 & 6.99\% & 13.13 & 42.17 & 10.36\% & 17.51 & 51.58 & 13.74\% & 21.94 & 59.57 & 17.32\% \\
         Time-LLM  & 17.59 & 29.34 & 16.81\% & 34.42 & 57.90 & 38.22\% & 29.41 & 49.57 & 31.21\% & 42.93 & 79.07 & 48.10\% \\
         GTS       & 9.04  & 29.91 & 6.76\% & 11.47 & 40.80 & \textcolor{red}{7.72\%} & \textcolor{blue}{14.42} & 47.81 & 11.53\% & 19.04 & 55.69 & \textcolor{blue}{14.62\%} \\
         GWN       & 10.21 & 30.34 & 6.40\% & 15.40 & 40.68 & 11.76\% & 18.00 & 49.89 & 14.54\% & 22.98 & 55.67 & 18.41\%\\
         ST-GFSL   & 9.47  & 30.12 & 6.93\% & 12.38 & 41.24 & 9.92\% & 16.31 & 49.55 & 12.84\% & 20.35 & 56.28 & 16.17\% \\
         TPB       & 8.68  & \textcolor{blue}{29.01} & \textcolor{blue}{6.30\%} & 11.45 & \textcolor{blue}{39.99} & 9.06\% & 15.31 & 48.02 & 11.85\% & 19.31 & 54.87 & 15.01\% \\
         TransGTR  & \textcolor{blue}{8.14} & \textcolor{red}{28.47} & 6.56\% & \textcolor{blue}{10.91} & \textcolor{red}{39.36} & 8.30\% & 14.84 & \textcolor{blue}{47.48} & \textcolor{blue}{11.17\%} & \textcolor{blue}{18.77} & \textcolor{blue}{54.34} & 14.71\%\\
         \hline 
         STRATA-TS   & \textcolor{red}{7.09} & 32.67 & \textcolor{red}{4.73\%} & \textcolor{red}{9.97} & 40.45 & \textcolor{blue}{7.76\%} & \textcolor{red}{13.26} & \textcolor{red}{46.94} & \textcolor{red}{10.36\%} & \textcolor{red}{16.61} & \textcolor{red}{52.16} & \textcolor{red}{12.73\%}\\
         \hline
         \multirow{2}*{\rotatebox{90}{ }}&  & 75 mins &  &  & 90 mins &  &  & 105 mins & & & 120 mins & \\ 
        &MAE& RMSE & MAPE &MAE& RMSE & MAPE&MAE& RMSE & MAPE&MAE& RMSE & MAPE\\ 
        \hline 
         DLinear   & 32.38 & 68.90 & 22.27\% & 38.08 & 76.46 & 26.61\% & 43.82 & 83.92 & 31.93\% & 50.09 & 91.28 & 37.35\% \\
         PatchTST  & 37.80 & 79.67 & 33.03\% & 43.83 & 87.85 & 37.91\% & 49.26 & 95.35 & 43.47\% & 53.65 & 101.95 & 48.26\% \\
         GPT4TS    & 26.68 & 67.25 & 21.81\% & 31.50 & 74.13 & 26.01\% & 36.54 & 80.98 & 31.35\% & 41.47 & 87.82 & 36.01\% \\       
         Time-LLM  & 39.27 & 78.49 & 37.91\% & 41.24 & 82.04 & 38.39\% & 45.22 & 87.76 & 42.21\% & 59.49 & 97.07 & 53.47\% \\
         GTS       & 24.39 & 62.49 & 19.33\% & 29.53 & 67.86 & 22.37\% & 34.70 & 71.74 & \textcolor{blue}{25.34\%} & 39.90 & 79.57 & \textcolor{blue}{28.87\%} \\
         GWN       & 26.30 & 61.51 & 22.97\% & 30.97 & 66.82 & 29.03\% & 33.05 & 72.09 & 31.78\% & 37.31 & 78.21 & 37.82\%\\
         ST-GFSL   & 24.19 & 62.43 & 18.65\% & 28.55 & 67.32 & 22.17\% & 32.29 & 71.95 & 25.70\% & 35.59 & 76.39 & 29.31\% \\
         TPB       & 23.97 & 62.20 & \textcolor{blue}{18.44\%} & 28.32 & 67.10 & \textcolor{blue}{21.96\%} & 32.06 & 71.70 & 25.48\% & 35.33 & 76.15 & 29.09\% \\
         TransGTR  & \textcolor{blue}{22.77} & \textcolor{blue}{60.81} & 18.50\% & \textcolor{blue}{26.67} & \textcolor{blue}{65.66} & 22.11\% & \textcolor{blue}{30.26} & \textcolor{blue}{70.40} & 25.92\% & \textcolor{blue}{33.35} & \textcolor{blue}{74.91} & 30.09\%\\
         \hline 
         STRATA-TS   & \textcolor{red}{20.44} & \textcolor{red}{59.30} & \textcolor{red}{16.36\%} & \textcolor{red}{24.08} & \textcolor{red}{64.99} & \textcolor{red}{19.15\%} & \textcolor{red}{27.45} & \textcolor{red}{68.55} & \textcolor{red}{22.12\%} & \textcolor{red}{30.94} & \textcolor{red}{72.74} & \textcolor{red}{23.63\%}\\
         \hline
         \multirow{2}*{\rotatebox{90}{ }}&  & 135 mins &  &  & 150 mins &  &  & 165 mins & & & 180 mins & \\ 
        &MAE& RMSE & MAPE &MAE& RMSE & MAPE&MAE& RMSE & MAPE&MAE& RMSE & MAPE\\ 
        \hline 
         DLinear   & 54.72 & 98.37 & 43.53\% & 59.69 & 105.27 & 49.13\% & 65.91 & 111.92 & 54.65\% & 72.70 & 118.43 & 60.12\% \\
         PatchTST  & 59.17 & 109.30 & 54.85\% & 64.85 & 116.68 & 59.77\% & 70.41 & 123.70 & 66.21\% & 75.48 & 130.20 & 73.00\% \\
         GPT4TS    & 46.79 & 94.71 & 42.82\% & 51.83 & 101.43 & 48.25\% & 56.88 & 108.02 & 53.84\% & 61.69 & 114.37 & 59.03\% \\
         Time-LLM  & 62.61 & 98.56 & 53.97\% & 60.77 & 106.40 & 52.28\% & 65.29 & 115.59 & 46.47\% & 66.16 & 117.74 & 46.12\% \\
         GTS       & 45.08 & 87.17 & \textcolor{blue}{32.77\%} & 49.79 & 93.83 & \textcolor{blue}{37.65\%} & 54.46 & 100.19 & \textcolor{blue}{42.79\%} & 56.37 & 112.23 & 53.37\% \\
         GWN       & 40.89 & 82.02 & 41.47\% & 44.46 & 87.74 & 47.25\% & 48.03 & 91.86 & 51.69\% & 51.10 & 95.30 & 54.79\%\\
         ST-GFSL   & 38.29 & 81.43 & 35.91\% & 41.83 & 86.12 & 40.11\% & 45.35 & 90.64 & 43.82\% & 48.37 & 95.13 & 46.27\% \\
         TPB       & 38.04 & 81.14 & 35.65\% & 41.58 & 85.83 & 39.84\% & 45.08 & 90.32 & 43.54\% & 48.08 & 94.82 & 45.98\% \\
         TransGTR  & \textcolor{blue}{37.20} & \textcolor{blue}{79.79} & 35.71\% & \textcolor{blue}{40.38} & \textcolor{blue}{84.36} & 39.69\% & \textcolor{blue}{43.78} & \textcolor{blue}{88.95} & 43.02\% & \textcolor{blue}{46.61} & \textcolor{red}{93.28} & \textcolor{blue}{44.73\%}\\
         \hline 
         STRATA-TS   & \textcolor{red}{34.39} & \textcolor{red}{78.73} & \textcolor{red}{26.84\%} & \textcolor{red}{37.66} & \textcolor{red}{83.45} & \textcolor{red}{29.89\%} & \textcolor{red}{40.92} & \textcolor{red}{88.30} & \textcolor{red}{32.89\%} & \textcolor{red}{44.08} & \textcolor{blue}{93.63} & \textcolor{red}{35.16\%}\\
         \hline
    \end{tabular}
\end{table*}

\section{Experiment}\label{experiment}
In this section, we evaluate the effectiveness of the proposed STRATA-TS framework for cross-city time series forecasting. We begin by detailing the experimental setup, including dataset descriptions, evaluation protocols, and baseline comparisons. To validate the advantages of STRATA-TS, we conduct a comprehensive performance comparison against state-of-the-art forecasting and transfer learning methods. We further design ablation experiments to analyze the contribution of each component in STRATA-TS, such as the retriever component and the LLM reasoning enhancement component. Additionally, we demonstrate that STRATA-TS can serve as a plug-and-play enhancement to existing forecasting backbones by selectively incorporating useful source knowledge, improving transferability and robustness in data-scarce target domains.

\subsection{Experimental Setup}

\textbf{Datasets.} We conduct experiments on three real-world parking availability datasets to evaluate the cross-city forecasting performance of our proposed STRATA-TS framework. Specifically, we use the Singapore Carpark Dataset as the source domain and two datasets, Nottingham and Glasgow, as the target domains. The detailed statistics of all datasets are provided in Table \ref{tab:Datasetss}. The Singapore Carpark Dataset contains parking availability records from 139 carparks across Singapore, recorded at 15-minute intervals. This large-scale dataset serves as the source domain due to its data abundance and high spatial coverage. The Nottingham Dataset includes 15-minute interval records from 11 carparks in Nottingham, UK, and was collected from the official Tramlink Nottingham website. The Glasgow Dataset consists of data from 7 carparks in Glasgow, UK, also recorded at 15-minute intervals. All datasets are split into training, validation, and testing sets using a standard 70\%/10\%/20\% ratio based on chronological order. To emulate a transfer learning scenario, we restrict the training data of the target cities (Nottingham and Glasgow) to only the most recent three days available within their respective training splits, following a similar setup to \cite{ST-GFS}. This setup allows us to rigorously evaluate the model’s ability to transfer knowledge from a data-rich source city to target cities with limited training data. We follow GTS~\cite{GTS} and use fixed input and output windows of $L_{\text{in}}=L_{\text{out}}=12$ steps (i.e., three hours of context predicting the subsequent three hours).

\textbf{Baselines.} We benchmark ten representative forecasting methods spanning statistical models, spatio–temporal graph neural networks (STGNNs), general time-series learners, and cross-city transfer approaches: \emph{ARIMA}\cite{ARIMA} (classical statistical baseline); \emph{Graph WaveNet}\cite{GRAPHWaveNet} and \emph{GTS}\cite{GTS} (STGNNs); \emph{DLinear}\cite{dlinear}, \emph{PatchTST}\cite{patchtst}, \emph{GPT4TS}\cite{GPT4TS}, and \emph{TimeLLM}\cite{Timellm} (widely used sequence-modeling baselines); and \emph{ST-GFSL}\cite{ST-GFS}, \emph{TPB}\cite{TPB} and \emph{TransGTR}\cite{TransGTR} (cross-city transfer baselines). We evaluate model performance using three standard metrics commonly employed in multivariate time series forecasting: Mean Absolute Error (MAE), Root Mean Squared Error (RMSE), and Mean Absolute Percentage Error (MAPE). Lower values across these metrics indicate better predictive accuracy.

\begin{itemize}
    \item \textbf{DLinear}~\cite{dlinear}: A decomposition-based linear model that splits time series into trend and seasonal components, providing a strong and efficient baseline for long-term forecasting.
    \item \textbf{PatchTST}~\cite{patchtst}: A transformer-based model that tokenizes time series into patches, allowing self-attention to capture both local and global temporal dependencies.
    \item \textbf{GPT4TS}~\cite{GPT4TS}: A large language model adapted for time series forecasting by reprogramming numerical inputs into tokenized sequences, enabling autoregressive prediction.
    \item \textbf{Time-LLM}~\cite{Timellm}: A framework that reprograms temporal data into prompt-based inputs for pretrained LLMs, leveraging their reasoning ability to improve sequence forecasting.
    \item \textbf{GTS}~\cite{GTS}: A graph-based forecasting framework that jointly learns temporal dynamics and the underlying graph structure to capture spatial correlations adaptively.
    \item \textbf{GraphWaveNet (GWN)}~\cite{GRAPHWaveNet}: A spatio-temporal deep learning model that employs dilated causal convolution and adaptive adjacency matrices to model complex dependencies.
    \item \textbf{ST-GFSL}~\cite{ST-GFS}: A spatio-temporal graph few-shot learning framework that utilizes meta-learning to transfer knowledge across cities with limited training data.
    \item \textbf{TPB}~\cite{TPB}: A cross-city transfer learning method that re-weights source city regions according to their relevance to the target city, reducing negative transfer effects.
    \item \textbf{TransGTR}~\cite{TransGTR}: A transferable graph structure learning framework that adapts source-domain graph construction to target domains, enabling robust cross-city forecasting.
\end{itemize}

\begin{table*}[!htbp] 
    \centering  
    \caption{120 mins Performance comparison on Glasgow dataset trained on SG dataset. Best results in \textcolor{red}{red}, second best in \textcolor{blue}{blue}.}\label{tab:glasgow}
    \begin{tabular}{c|c c c|c c c|c c c|c c c}  
        \hline 
          \multirow{2}*{\rotatebox{90}{ }}&  & 15 mins &  &  & 30 mins &  &  & 45 mins & & & 60 mins & \\ 
        &MAE& RMSE & MAPE &MAE& RMSE & MAPE&MAE& RMSE & MAPE&MAE& RMSE & MAPE\\ 
        \hline 
         DLinear   & 10.62 & 34.06 & \textcolor{blue}{1.89\%} & 15.47 & 49.12 & 2.93\% & 19.99 & 61.26 & 3.94\% & 27.11 & 72.85 & 5.33\% \\
         PatchTST  & 20.52 & 55.05 & 3.95\% & 27.61 & 68.64 & 5.34\% & 31.04 & 77.74 & 6.15\% & 37.98 & 89.32 & 7.55\% \\
         GPT4TS    & 9.93  & 34.82 & 1.98\% & 13.99 & 48.83 & 2.78\% & 18.52 & 60.56 & 3.67\% & 23.43 & 71.21 & 4.66\% \\    
         TimeLLM   & 10.32 & 35.78 & 2.12\% &14.29 & 50.21 & 3.03\% & 20.19 & 63.11& 4.73\%& 26.82& 75.94 &5.12\%\\
         GTS       & 10.28 & 34.38 & 2.76\% & \textcolor{blue}{13.62} & \textcolor{red}{41.38} & 3.25\% & \textcolor{blue}{16.72} & 54.98 & 3.73\% & \textcolor{blue}{21.25} & 61.71 & 4.44\% \\
         GWN       & 12.93 & 36.30 & 2.35\% & 20.63 & 53.76 & 3.34\% & 23.33 & 59.61 & 3.73\% & 26.88 & 62.41 & 4.52\% \\
         ST-GFSL   & 10.48 & 33.01 & 2.11\% & 14.45 & 45.53 & 2.97\% & 18.36 & 54.70 & 3.67\% & 22.81 & 63.49 & 4.41\% \\
         TPB       & 10.25 & 32.62 & 2.01\% & 14.22 & 45.02 & 2.85\% & 18.11 & 54.19 & 3.52\% & 22.43 & 63.05 & \textcolor{blue}{4.20\%} \\
         TransGTR  & \textcolor{blue}{9.84} & \textcolor{blue}{32.05} & 1.96\% & 13.67 & 44.36 & \textcolor{blue}{2.65\%} & 17.03 & \textcolor{blue}{52.45} & \textcolor{blue}{3.17\%} & 21.90 & \textcolor{blue}{61.33} & 4.23\% \\
         \hline 
         STRATA-TS   & \textcolor{red}{7.50} & \textcolor{red}{31.95} & \textcolor{red}{1.50\%} & \textcolor{red}{11.78} & \textcolor{blue}{43.60} & \textcolor{red}{2.34\%} & \textcolor{red}{15.80} & \textcolor{red}{51.96} & \textcolor{red}{3.10\%} & \textcolor{red}{19.44} & \textcolor{red}{60.05} & \textcolor{red}{3.84\%} \\
         \hline
         \multirow{2}*{\rotatebox{90}{ }}&  & 75 mins &  &  & 90 mins &  &  & 105 mins & & & 120 mins & \\ 
        &MAE& RMSE & MAPE &MAE& RMSE & MAPE&MAE& RMSE & MAPE&MAE& RMSE & MAPE\\ 
        \hline 
         DLinear   & 32.92 & 83.82 & 6.58\% & 38.32 & 94.11 & 7.78\% & 44.88 & 103.91 & 9.12\% & 51.69 & 113.30 & 10.50\% \\
         PatchTST  & 44.69 & 99.38 & 8.91\% & 50.67 & 109.74 & 10.30\% & 56.79 & 118.77 & 11.48\% & 62.73 & 127.43 & 12.71\% \\
         GPT4TS    & 29.06 & 81.29 & 5.83\% & 34.55 & 90.64 & 6.98\% & 40.05 & 99.60 & 8.15\% & 45.96 & 108.44 & 9.42\% \\
         TimeLLM   & 31.24 & 85.93 & 6.63\% &36.12 & 94.08 & 7.81\% & 42.12 &104.54 & 10.09\%&49.21 & 110.78 &10.98\%\\
         GTS       & 26.34 & 69.60 & 5.24\% & 31.20 & 76.24 & 6.06\% & 35.73 & 80.86 & 6.81\% & 40.13 & \textcolor{blue}{85.38} & 7.58\% \\
         GWN       & 29.74 & 68.01 & 5.30\% & 32.48 & 75.16 & 6.08\% & 35.71 & 80.20 & 6.78\% & 40.18 & 85.88 & 7.77\% \\
         ST-GFSL   & 26.34 & 68.80 & 5.22\% & 30.22 & 75.33 & 5.96\% & 33.81 & 80.38 & 6.53\% & 37.63 & 86.01 & 7.39\% \\
         TPB       & 26.07 & 68.41 & 5.08\% & 29.95 & 75.04 & 5.77\% & 33.52 & 79.99 & 6.29\% & 37.30 & 85.64 & 7.15\% \\
         TransGTR  & \textcolor{blue}{25.53} & \textcolor{blue}{67.60} & \textcolor{blue}{4.89\%} & \textcolor{blue}{29.47} & \textcolor{blue}{74.52} & \textcolor{blue}{5.73\%} & \textcolor{blue}{32.22} & \textcolor{blue}{79.32} & \textcolor{blue}{6.27\%} & \textcolor{blue}{36.14} & 85.74 & \textcolor{blue}{7.09\%} \\
         \hline 
         STRATA-TS   & \textcolor{red}{23.41} & \textcolor{red}{66.04} & \textcolor{red}{4.64\%} & \textcolor{red}{27.57} & \textcolor{red}{73.99} & \textcolor{red}{5.47\%} & \textcolor{red}{30.23} & \textcolor{red}{77.20} & \textcolor{red}{6.01\%} & \textcolor{red}{34.25} & \textcolor{red}{84.28} & \textcolor{red}{6.81\%} \\
         \hline
         \multirow{2}*{\rotatebox{90}{ }}&  & 135 mins &  &  & 150 mins &  &  & 165 mins & & & 180 mins & \\ 
        &MAE& RMSE & MAPE &MAE& RMSE & MAPE&MAE& RMSE & MAPE&MAE& RMSE & MAPE\\ 
        \hline 
         DLinear   & 57.13 & 122.56 & 11.73\% & 66.49 & 130.64 & 13.34\% & 75.43 & 138.96 & 14.90\% & 81.83 & 147.10 & 16.18\% \\
         PatchTST  & 68.86 & 136.35 & 13.92\% & 76.05 & 146.74 & 15.82\% & 81.97 & 153.37 & 16.50\% & 86.77 & 160.69 & 17.85\% \\
         GPT4TS    & 51.91 & 117.11 & 10.71\% & 57.76 & 125.48 & 11.98\% & 63.59 & 133.64 & 13.24\% & 69.35 & 141.49 & 14.50\% \\
         TimeLLM   & 55.21 & 120.24 & 12.54\% & 60.84 & 129.21 & 13.24\% & 67.92  &137.31 & 14.21\%&75.75 & 148.21 & 16.50\%\\
         GTS       & 44.61 & 92.96 & 8.41\% & 48.72 & 97.36 & 9.24\% & 52.61 & 101.52 & 10.01\% & 58.35 & 114.80 & 10.21\% \\
         GWN       & 42.38 & 92.24 & 8.22\% & 45.72 & 97.70 & 8.84\% & 49.37 & \textcolor{blue}{100.51} & 9.55\% & 52.09 & 105.49 & 10.14\% \\
         ST-GFSL   & 41.05 & 95.02 & 8.12\% & 44.72 & 99.88 & 8.83\% & 48.42 & 104.37 & 9.52\% & 51.73 & 108.91 & 10.28\% \\
         TPB       & 40.81 & 94.70 & 7.98\% & 44.45 & 99.52 & 8.63\% & 48.14 & 103.96 & \textcolor{blue}{9.33\%} & 51.41 & 108.45 & \textcolor{blue}{10.03\%} \\
         TransGTR  & \textcolor{blue}{39.51} & \textcolor{blue}{91.27} & \textcolor{blue}{7.86\%} & \textcolor{blue}{43.10} & \textcolor{blue}{96.14} & \textcolor{blue}{8.61\%} & \textcolor{blue}{46.93} & \textcolor{blue}{100.51} & \textcolor{blue}{9.33\%} & \textcolor{blue}{50.92} & \textcolor{red}{105.23} & 10.10\% \\
         \hline
         STRATA-TS   & \textcolor{red}{37.47} & \textcolor{red}{90.86} & \textcolor{red}{7.51\%} & \textcolor{red}{40.47} & \textcolor{red}{95.46} & \textcolor{red}{8.14\%} & \textcolor{red}{43.38} & \textcolor{red}{100.28} & \textcolor{red}{8.73\%} & \textcolor{red}{45.60} & \textcolor{blue}{105.99} & \textcolor{red}{9.15\%} \\
         \hline
    \end{tabular}
\end{table*}

\textbf{Implementation Details.} All experiments are conducted in PyTorch on a Linux workstation equipped with an Intel\textsuperscript{\textregistered} Core\texttrademark~i7-13700K CPU and a single NVIDIA A6000 GPU (48\,GB). We enable FP16 mixed-precision training and gradient checkpointing to reduce memory usage. The LLaMa-3-8B model is quantized to 4-bit~\cite{quanti} and adapted using LoRA~\cite{lora} with a rank of $r{=}64$. We use an initial learning rate of $2\times10^{-4}$ with the Adam optimizer. STRATA-TS is fine-tuned for 5~epochs using LLaMa-3-8B, while TransGTR~\cite{TransGTR} is pre-trained as the base model following its official default configuration.

\subsection{Performance of STRATA-TS}

We evaluate the performance of STRATA-TS against a diverse set of baselines and experimental results are shown in Table~\ref{tab:not} and Table~\ref{tab:glasgow}. Across all evaluations, STRATA-TS achieves the best performance on \textbf{32 out of 36} evaluations for Nottingham and \textbf{34 out of 36} for Glasgow, yielding a total of \textbf{66 out of 72} wins. This consistent dominance over strong baselines such as TransGTR, GTS, and GPT4TS underscores the robustness of our retrieval-augmented reasoning approach.

\paragraph{Nottingham results.}
On the Nottingham dataset (Table~\ref{tab:not}), STRATA-TS consistently outperforms all baselines across most horizons. At a 15-minute horizon, it achieves an MAE of 7.09 and MAPE of 4.73\%, representing a 12.9\% reduction in MAE and a 27.9\% reduction in MAPE compared to TransGTR. Even at the longest horizon (180 minutes), STRATA-TS maintains clear superiority with an MAE of 44.08 and MAPE of 35.16\%, outperforming TransGTR by 5.4\% and 21.4\%, respectively. These results highlight its ability to mitigate long-term error propagation—a key limitation of conventional transfer learning models.

\paragraph{Glasgow results.}
A similar trend is observed for the Glasgow dataset (Table~\ref{tab:glasgow}). At the shortest horizon (15 minutes), STRATA-TS records an MAE of 7.50 and MAPE of 1.50\%, improving over TransGTR by 23.8\% and 23.5\%, respectively. At 180 minutes, it achieves an MAE of 45.60 and MAPE of 9.15\%, outperforming TransGTR by 10.5\% and 9.4\%. Gains are especially pronounced at intermediate horizons (75--135 minutes), where STRATA-TS demonstrates greater resilience to temporal degradation compared to all baselines.

Three consistent patterns emerge across both datasets:  
(1) \textbf{Consistent superiority over baselines:} STRATA-TS dominates the majority of evaluations, with particularly strong margins at longer horizons.  
(2) \textbf{Reduced long-horizon degradation:} While baseline models exhibit steep error growth with increasing horizons, STRATA-TS displays a slower error growth rate, indicating that retrieval-augmented LLM inference effectively constrains error accumulation.  
(3) \textbf{Synergy between retrieval and reasoning:} The performance gap between STRATA-TS and pure ST models (e.g., TransGTR, GTS) shows that integrating semantically relevant retrieved sequences with causal reasoning boosts predictive accuracy beyond what parameter transfer alone can achieve. By combining selective retrieval of domain-relevant sequences with LLM-driven inference, STRATA-TS effectively leverages transferable knowledge while avoiding negative transfer effects. This enables superior accuracy across both short- and long-term horizons, establishing STRATA-TS as a robust and generalizable solution for cross-city time series forecasting.

\subsection{Ablation Study}
\label{sec:ablation}

\begin{figure}[t]
    \centering
    \includegraphics[width=0.98\linewidth]{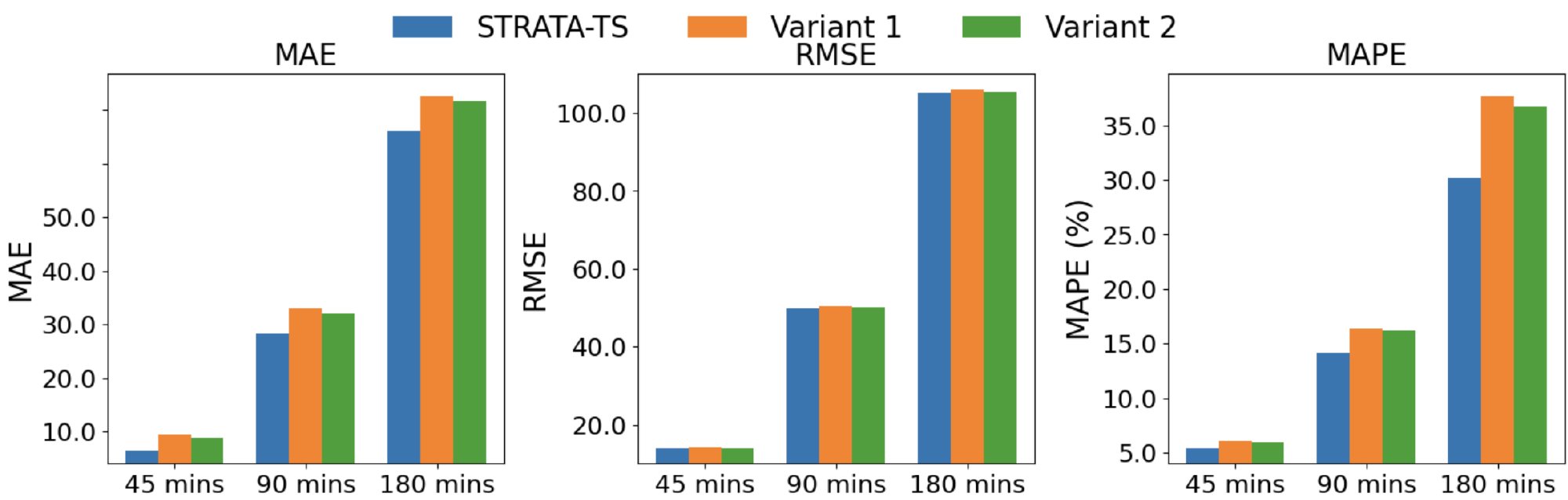}
    \caption{Ablation study results on the Nottingham dataset. Removing either the semantic retriever (\emph{Random-Centroid}) or the powerful reasoning LLM (\emph{w/ LLaMA-3-8B}) leads to a performance drop in MAE and MAPE, while RMSE changes are relatively smaller.}
    \label{fig:ablation_bar}
\end{figure}

To assess the contribution of each core component in \textbf{STRATA-TS}, we design two controlled variants of the model on the Nottingham dataset, each modifying a single module while keeping all other settings identical to the original. The performance of these variants is compared to the full \textbf{STRATA-TS} system, and the results are visualized in Fig.~\ref{fig:ablation_bar}.

\paragraph{Variant 1: Random-Centroid Retrieval.}  
In this variant, we study the role of the retriever module by replacing the similarity-based retrieval mechanism with a \emph{k}-means clustering over the source-city embeddings. For each cluster, a single centroid sequence and its associated textual description are randomly selected as the long-term context for all target queries assigned to that cluster. This process removes fine-grained semantic matching between source and target sequences, potentially degrading the quality of retrieved patterns.

\paragraph{Variant 2: Open-Source LLM Inference.}  
In this variant, we evaluate the importance of the reasoning capacity in the Retrieval-Guided Reasoning Module by replacing the proprietary LLMs (GPT-o1, DeepSeek-R1) with a directly deployed open-source \textbf{LLaMA-3-8B} model. The same input format is preserved, but without the distilled high-quality reasoning traces from GPT-o1, the reasoning process is expected to be less nuanced.

\paragraph{Analysis.}  
Both ablation settings result in a noticeable drop in forecasting accuracy, particularly in MAE and MAPE, while RMSE changes are relatively smaller. This indicates that these components are especially effective for improving forecasting performance, whereas RMSE—being more sensitive to rare large deviations—is less affected.  
For Variant 1, the performance drop confirms that semantic retrieval from the source domain is critical for providing representative and domain-relevant long-term patterns, which directly enhance the downstream LLM reasoning stage.  
For Variant 2, the weaker performance shows that while LLaMA-3-8B can perform some correction, it lacks the advanced cross-city causal inference capabilities of GPT-o1, validating the effectiveness of our two-stage reasoning–distillation approach.

\subsection{Visualization}
\begin{figure}[t]
    \centering
    \includegraphics[width=0.98\linewidth]{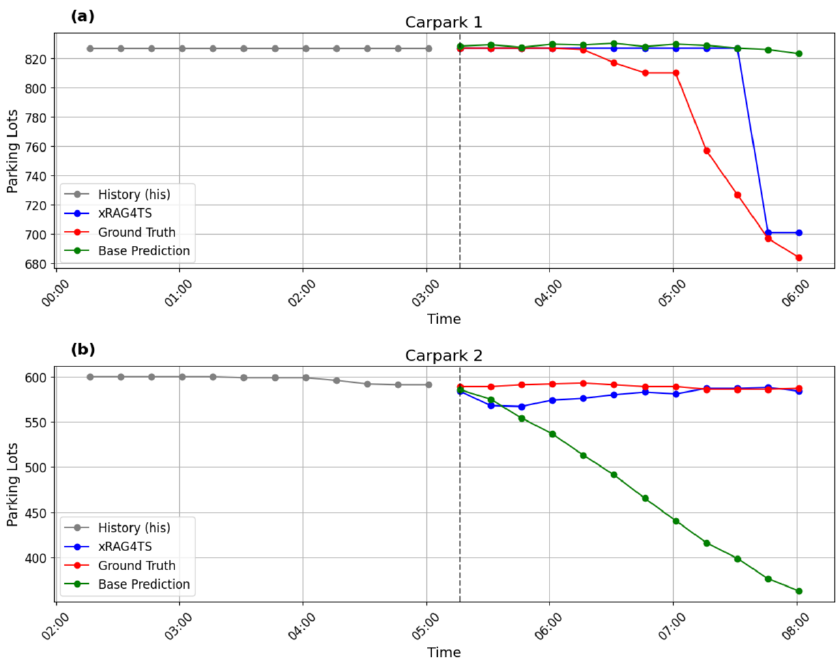}
    \caption{Visualization of forecasting results. (a) For Carpark~1, \textsc{STRATA-TS} accurately captures the decline in parking lot availability and follows the ground truth closely. (b) Compared with the base model TransGTR, \textsc{STRATA-TS} correctly identifies that the parking availability remains stable, avoiding the erroneous sharp decrease predicted by TransGTR. These examples illustrate the effectiveness and robustness of our proposed framework.}
    \label{fig:vis}
\end{figure}

To provide an intuitive understanding of the forecasting behavior of our proposed framework, we present visualization results in Fig.~\ref{fig:vis}. As shown in Fig.~\ref{fig:vis}(a), for Carpark~1, \textsc{STRATA-TS} is able to recognize the decline in parking lot availability and closely follow the ground-truth trend, demonstrating its ability to capture dynamic variations. In contrast, Fig.~\ref{fig:vis}(b) highlights a scenario where the baseline model TransGTR incorrectly predicts a sharp decrease in availability, while \textsc{STRATA-TS} successfully identifies that the parking availability remains relatively stable. These visualizations clearly indicate that \textsc{STRATA-TS} not only adapts to diverse temporal dynamics but also avoids overestimating drastic fluctuations, thereby producing more reliable and accurate forecasts. Overall, these case studies confirm the effectiveness of our retrieval-augmented design in improving forecasting robustness and fidelity.

\begin{figure}[t]
    \centering
    \includegraphics[width=0.98\linewidth]{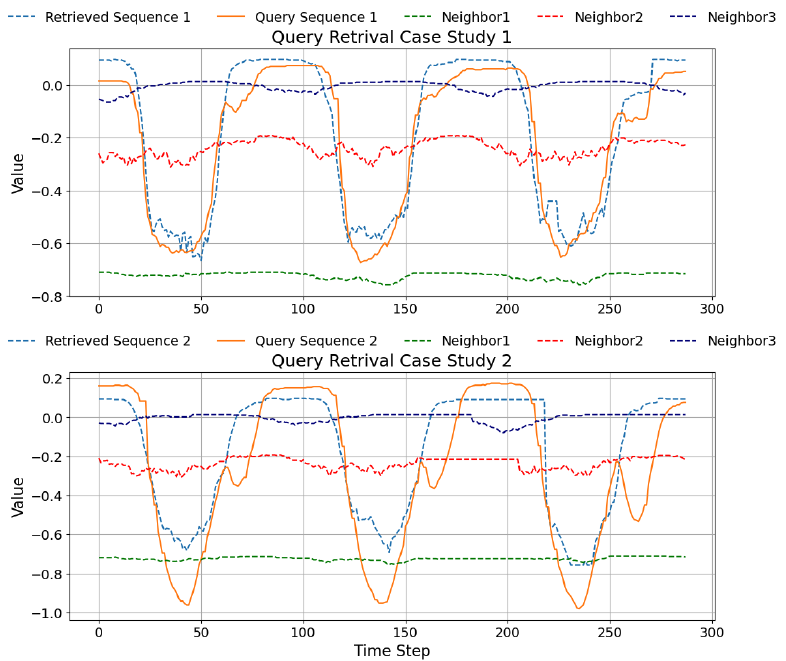}
    \caption{Case studies demonstrating the effectiveness of the retriever module. In both the top and bottom examples, the retrieved source sequence (after $z$-normalization) matches the target query sequence closely, while neighboring sequences deviate more noticeably. This confirms that the retriever accurately identifies semantically relevant patterns to support robust cross-city forecasting.}
    \label{fig:retriever}
\end{figure}

To further validate the effectiveness of our selective retriever, we illustrate two representative case studies in Fig.~\ref{fig:retriever}. In each case, the top and bottom panels correspond to different query sequences from the target domain. After applying $z$-normalization, the retrieved source sequence shows a strong alignment with the query sequence, whereas the neighboring sequences deviate more significantly. This demonstrates that our retriever is capable of capturing semantically relevant and temporally consistent patterns from the source city, ensuring that the most informative knowledge is transferred to support forecasting in the target city.

\section{Conclusion}

In this work, we present STRATA-TS, a retrieval-augmented framework for cross-city time series forecasting that selectively transfers relevant knowledge from the data-rich source cities to improve forecasting accuracy in data-scarce target cities. By combining a Target-Aware Retriever Module, STRATA-TS filters out irrelevant knowledge and enhances transfer effectiveness. Additionally, we incorporate an Retrieval-Guided Reasoning Module, where GPT-o1’s reasoning capability is adapted into a compact LLaMA-3-8B model via supervised fine-tuning. Our method is plug-and-play compatible with existing transfer frameworks, such as TransGTR, and significantly boosts their performance. Extensive experiments on real-world parking availability datasets across multiple cities validate the effectiveness and generalizability of STRATA-TS. Results show that our framework consistently outperforms existing baselines, highlighting the potential of retrieval-based transfer and LLM-driven forecasting in spatio-temporal domains. Future work includes extending STRATA-TS to multi-modal urban data sources from multi-domains and investigating scalable retrieval mechanisms for larger temporal corpora.



\bibliographystyle{ACM-Reference-Format}
\bibliography{main}

\end{document}